\documentclass[conference]{IEEEtran}
\IEEEoverridecommandlockouts

\usepackage{cite}
\usepackage{amsmath,amssymb,amsfonts}
\usepackage{algorithmic}
\usepackage{graphicx}
\usepackage{textcomp}
\usepackage{xcolor}
\usepackage{multirow}
\usepackage[hidelinks]{hyperref}
\usepackage{bbm}
\def\BibTeX{{\rm B\kern-.05em{\sc i\kern-.025em b}\kern-.08em
    T\kern-.1667em\lower.7ex\hbox{E}\kern-.125emX}}
\begin{document}

\title{Accelerating SNN Training with\\Stochastic Parallelizable Spiking Neurons}

\author{
\IEEEauthorblockN{Sidi Yaya Arnaud Yarga}
\IEEEauthorblockA{
\textit{Department of Electrical and Computer Engineering} \\
\textit{Université de Sherbrooke}, 
Sherbrooke, QC, Canada \\
Email: sidi.yaya.arnaud.yarga@usherbrooke.ca}
\and
\IEEEauthorblockN{Sean U. N. Wood}
\IEEEauthorblockA{
\textit{Department of Electrical and Computer Engineering} \\
\textit{Université de Sherbrooke}, 
Sherbrooke, QC, Canada \\
Email: sean.wood@usherbrooke.ca}
}

\newcommand{\todo}[1]{\textcolor{red}{#1}}
\newcommand{\newtext}[1]{\textcolor{blue}{#1}}
\newcommand{\warningtext}[1]{\textcolor{orange}{#1}}

\maketitle

\begin{abstract}

Spiking neural networks (SNN) are able to learn spatiotemporal features while using less energy, especially on neuromorphic hardware. The most widely used spiking neuron in deep learning is the Leaky Integrate and Fire (LIF) neuron. LIF neurons operate sequentially, however, since the computation of state at time \(t\) relies on the state at time \(t-1\) being computed. This limitation is shared with Recurrent Neural Networks (RNN) and results in slow training on Graphics Processing Units (GPU). In this paper, we propose the Stochastic Parallelizable Spiking Neuron (SPSN) to overcome the sequential training limitation of LIF neurons. By separating the linear integration component from the non-linear spiking function, SPSN can be run in parallel over time. The proposed approach results in performance comparable with the state-of-the-art for feedforward neural networks on the Spiking Heidelberg Digits (SHD) dataset, outperforming LIF networks while training 10 times faster and outperforming non-spiking networks with the same network architecture. For longer input sequences of 10\,000 time-steps, we show that the proposed approach results in 4000 times faster training, thus demonstrating the potential of the proposed approach to accelerate SNN training for very large datasets.
\end{abstract}

\begin{IEEEkeywords}
spiking neural networks, neuromorphic computing, stochastic neurons, parallelization, hardware acceleration
\end{IEEEkeywords}

\section{Introduction}
Graphics Processing Units (GPU) \cite{owens2008gpu,nickolls2010gpu} and Tensor Processing Units (TPU) \cite{jouppi2017datacenter} have been fundamental to the success of deep learning. The parallelization capacities provided by these devices allow deep neural network training to be accelerated dramatically \cite{fan2004gpu, sanders2010cuda, wang2019benchmarking}. This speed up is critical for training networks on large data sets such as ImageNet containing over 14 millions images \cite{deng2009imagenet} or the ``Colossal Clean Crawled Corpus" comprising 750 GB of clean and natural English text \cite{raffel2020exploring}. The performance achieved on such large datasets allowed the creation of real world applications in a wide variety of domains, leading to the deep learning revolution.

Spiking Neural Networks (SNN) are dynamical systems that evolve over time \cite{ghosh2009spiking,tavanaei2019deep}. As with biological neurons, information must inherently be processed sequentially, thus preventing SNNs from taking full advantage of GPU and TPU hardware acceleration. This issue is common to recurrent neural networks (RNN) as well, where the computation of the network state at time \(t\) depends on the network state at time \(t-1\). A few approaches have been proposed in the literature to accelerate RNN training on GPUs \cite{li2014large}. For example, an approach based on batch bucketing and data parallelization on multiple GPUs was proposed in \cite{khomenko2016accelerating} resulting in an acceleration factor of 4. More recently, the authors in \cite{voelker2019legendre} proposed the Legendre Memory Unit (LMU) architecture that outperforms other RNN architectures including the Long short-term memory (LSTM) and Gated Recurrent Unit (GRU) on a wide variety of tasks. In \cite{chilkuri2021parallelizing}, the authors subsequently leveraged the fact that the LMU's linear time-invariant (LTI) system state update equation can be written in a non-sequential fashion \cite{aastrom2021feedback}. This allowed for the parallel computation of the hidden state of the LTI system over time, thus overcoming the sequential training limitation of RNNs trained on GPUs. Parallelization over time of the LMU led to training times up to 200 times faster.

In the deep learning literature, spiking neurons generally produce spikes according to deterministic rules. However, biological data suggest that computation in the brain is stochastic in nature. If one tries to induce the brain to carry out the same computation again and again, the spiking activity varies substantially from trial to trial \cite{maass2015spike}. This variability of observed neural responses is believed to be caused by stochastic signal transmission at synapses and stochastic local amplifiers in the dendrites of neurons \cite{maass2015spike}.
It has also been suggested that stochastic neuronal output may result from high-frequency excitatory and inhibitory Poisson stimuli \cite{dold2019stochasticity}.

Despite their inherent uncertainty, stochastic SNNs have been used to solve real problems in the literature. For example, in \cite{fonseca2017using} a stochastic SNN implemented on the SpiNNaker hardware \cite{furber2014spinnaker} is used to solve Constraint Satisfaction Problems (CSP). The solver works as a stochastic dynamical system that is attracted by the configuration that solves the CSP. As shown by authors,  the presence of noise allows an optimal exploration of the space of configurations, seeking the satisfiability of all the constraints. 
In \cite{buesing2011neural}, the authors propose a neural network model whose neural activity is  demonstrated theoretically to implement Markov chain Monte Carlo (MCMC) sampling of a given distribution.
Based on this demonstration, \cite{petrovici2013stochastic} extends Bayesian spiking network implementations to deterministic neuron models widely used in computational neuroscience.

In this paper, we propose the Stochastic Parallelizable Spiking Neuron (SPSN) to accelerate SNN training on GPUs\footnote{A Python implementation of the proposed SPSN approach is made available online at: \url{https://github.com/NECOTIS/Stochastic-Parallelizable-Spiking-Neuron-SPSN}}. The proposed approach separates the functioning of the traditional Leaky Integrate and Fire (LIF) neuron into a linear component that integrates the input and a stochastic non-linear component that generates spikes. The linear component is parallelized over time as an LTI system. The non-linear component is a stochastic firing process that is subsequently applied in parallel, independently at each point in time. 

This approach takes inspiration from the instantaneous firing probability function for approximating the experimentally observed stochastic spike generation of a biological neuron as proposed by \cite{maass2015spike}.
In addition to the SPSN model, we integrate data augmentation and spike frequency regularization methods to the training procedure to further improve classification performance and to reduce the amount of spikes required by the network during inference.

The manuscript is organized as follows. We begin with a comprehensive review of related literature Section \ref{sec:Related Work} followed by a development of the proposed SPSN model in Section \ref{sec:spiking-neurons}. Experimental procedures used to evaluate the model are outlined in Section \ref{sec:experiments} with the corresponding results reported in Section \ref{sec:results}. Section \ref{sec:discussion} presents the ensuing discussions and finally, Section \ref{sec:conclusion} provides the concluding remarks.

\section{Related Work}\label{sec:Related Work}
The need to accelerate SNN simulation has been widely addressed in the literature.
Numerous studies from the target fields of neuroscience and data-science have put forth effective solutions that leverage the advantages offered by GPUs.
However, we note that these simulations are still performed sequentially over time.

In \cite{brette2012simulating}, authors review the ongoing efforts towards simulating SNNs on GPU for neuroscientists. Various software packages such as NeMo \cite{fidjeland2009nemo}, GeNN \cite{yavuz2016genn} and Brian \cite{goodman2009brian} were investigated. Most of these packages decompose the SNN simulation into three main components: integrating the differential equations, propagating the spikes to target neurons and applying the effect of spikes on target neurons. On GPUs, parallelizing the numerical integration of differential equations is straightforward, because it follows the SIMD paradigm (Single Instruction, Multiple Data) \cite{brette2012simulating}. However, parallelizing spike propagation is more challenging, as it does not conform to the SIMD paradigm. Therefore, the main bottleneck for GPU simulation is the propagation of spikes across the network \cite{brette2012simulating}. To overcome this challenge, two main strategies have been proposed. The first strategy involves parallelizing over neurons, while the second strategy parallelizes over synaptic events, that is, over received spikes. In both strategies, GPU kernels need to access a large amount of memory at each timestep \cite{brette2012simulating}. Moreover, these strategies all run many unnecessary operation at each time-step, especially when spikes are relatively infrequent. \cite{kasap2018dynamic} then proposed a novel parallelization strategy, which utilizes dynamic parallelism for synaptic updating. This algorithm updates all post-synaptic currents for each action potential in parallel. As a result, they outperform the other two strategies when the spike occurrence is sparse in relation to the network size. By using a different approach, \cite{slazynski2012streaming} proposed the use of the Spike Response Model as the additive nature of membrane potential dynamics enables additional update parallelism. The authors further show that optimizing simulation algorithms and data structures to the GPU’s architecture has a significant pay-off \cite{slazynski2012streaming}.

On the data-science side, many software frameworks provide a set of ready-to-use tools that researchers can leverage. A review of 9 frameworks for the development of SNN that are specifically oriented towards machine learning has been done by \cite{manna2023frameworks}. It is shown that several of them like  Norse, PySNN, snnTorch, SpikingJelly and BindsNet use PyTorch as a base to leverage the GPU acceleration.

The common point of all previous works is that the simulations are performed sequentially over time. The proposed SPSN model, on the other hand, allows network simulation to be performed in parallel over time. A similar neuron model called the linear-nonlinear-Poisson model exists in the literature. However, it has been used primarily in neuroscience to characterize neural responses \cite{simoncelli2004characterization}.

\section{Spiking Neurons Models}\label{sec:spiking-neurons}
We begin by reviewing the Leaky Integrate and Fire (LIF) neuron on which the proposed Stochastic Parallelizable Spiking Neuron (SPSN) is based. We then develop the SPSN in detail including its parallel leaky integrator and stochastic firing components.

\begin{figure}[htbp]
\centerline{\includegraphics[width=1.0\columnwidth]{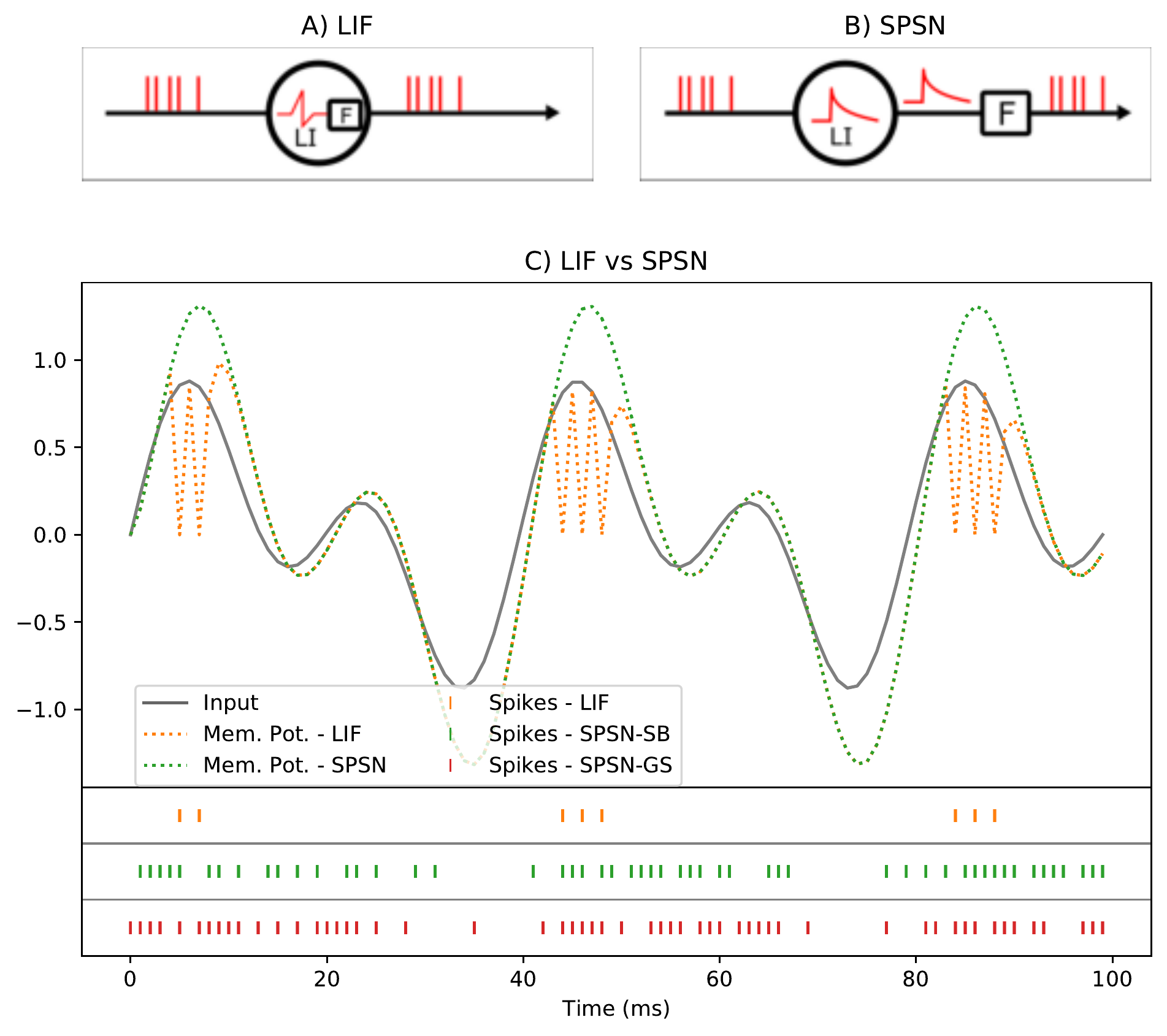}}
\caption{Illustration of the spiking neuron models: A) The LIF neuron: the spiking function is nested with the leaky integration process. B) The SPSN: the Leaky Integrating phase is separated from the spiking phase. C) Neuron outputs: spikes generated by the LIF neuron (orange) occur only when the membrane potential reaches the spiking threshold. For the SPSN (green and red) the density of spikes are correlated with the value of the membrane potential. Given the stochasticity of the firing process, spikes can even occur when the membrane potential is very low.}
\label{img:Architecture}
\end{figure}

\subsection{LIF Neuron}
Several spiking neuron models have been proposed in the literature that vary in biologically plausibility and computationally efficiency \cite{izhikevich2004model}. The most frequently used spiking neuron in the deep learning setting is the Leaky Integrate and Fire (LIF) neuron. The LIF neuron achieves fundamental neuro-computational features and requires less energy to be simulated \cite{izhikevich2004model}. It has been used in many contexts in the literature. For example, Intel used it to conceive the Loihi chip for modeling spiking neural networks \cite{davies2018loihi}. In \cite{hunsberger2015spiking}, authors used LIF neurons to classify CIFAR-10 and MNIST datasets while in \cite{corradi2019ecg} a cluster of LIF neurons was trained to distinguish 17 types of cardiac patterns trough electrocardiogram signals. A more detailed review of LIF neurons can be found in \cite{burkitt2006review}.

As depicted in Figure~\ref{img:Architecture}-A, the functioning of a LIF neuron can be separated into two phases: \emph{leaky integration} (LI) and \emph{firing} (F). Defined mathematically below, the leaky integration phase comprises the integration of the presynaptic input to the neuron current \eqref{eq:LIF_integ_i} and the integration of the current to the membrane potential \eqref{eq:LIF_integ_u}. The firing phase \eqref{eq:LIF_fire} occurs when the membrane potential reaches the spiking threshold, at which time a post-synaptic spike is emitted and the membrane potential is reset to the value \(u_r\) \eqref{eq:LIF_transition}. This process is illustrated in Figure~\ref{img:Architecture}-C and defined as follows,

\begin{equation}
 \frac{d\mathbf{i}}{dt}=-\frac{\mathbf{i}(t)}{\tau _{syn}}+\mathbf{x}(t)
 \label{eq:LIF_integ_i}
\end{equation}
\begin{equation}
 \tau _{mem}\frac{d\mathbf{u}}{dt} = -\mathbf{u}(t)+r\mathbf{i}(t)
 \label{eq:LIF_integ_u}
\end{equation}
\begin{equation}
 \mathbf{s}(t) = \Theta(\mathbf{u}(t)-u_{th})
 \label{eq:LIF_fire}
\end{equation}
\begin{equation}
 \mathbf{u}(t) = \mathbf{u}(t) (1-\mathbf{s}(t)) + u_r \mathbf{s}(t)
 \label{eq:LIF_transition}
\end{equation}
where \(\tau _{syn}\) is the synaptic time constant, \(\tau _{mem}\) is the membrane time constant, \(\mathbf{i}\) the input current, \(\mathbf{x}\) is the presynaptic input, \(\mathbf{u}\) is the membrane potential, \(r\) is the input resistance, \(\mathbf{s}\) defines the presence \((s(t)=1)\) or absence \((s(t)=0)\) of spikes, \(\Theta\) is the Heaviside function, \(u_{th}\) is the spiking threshold, and \(u_r\) is the membrane potential reset value. 

Using a small time-step \(\Delta_t\) in discrete time, \eqref{eq:LIF_integ_i} and \eqref{eq:LIF_integ_u} can be approximated using the Euler method. By fixing \(r=1\) and \(u_r=0\), \eqref{eq:LIF_integ_i} and \eqref{eq:LIF_integ_u} become respectively: 
\begin{equation}
 \mathbf{i}[n+1] = \alpha \mathbf{i}[n] + \mathbf{x}[n+1]
\label{eq:LIF_approx_i}
\end{equation}
\begin{equation}
 \mathbf{u}[n+1] =  \left ( \beta \mathbf{u}[n] + (1-\beta) \mathbf{i}[n+1] \right ) (1-\mathbf{s}[n])
\label{eq:LIF_approx_u}
\end{equation}
where \(\alpha=exp(-\Delta_t/\tau _{syn})\) and \(\beta=exp(-\Delta_t/\tau _{mem})\) are the decay strength of the input current and membrane potential respectively.

The derivative of the Heaviside function in \eqref{eq:LIF_fire} is the Dirac delta function, which equals to 0 almost everywhere and then prevents the direct use of backpropagation for training LIF-based neural networks. To overcome this, the surrogate gradient method proposed in \cite{neftci2019surrogate} is used.

\subsection{Stochastic Parallelizable Spiking Neuron (SPSN)}
The LIF neuron presented above is not parallelizable over time since in order to compute the membrane potential at time-step \(t\), we must first apply the non-linear firing function to the membrane potential at time-step \(t-1\). As we will see below, parallelization over time is only possible for linear systems. To overcome this limitation, we begin by modifying the functioning of the LIF neuron by removing the non-linear spiking function from the recurrence. With no reset applied to the membrane potential (see Figure~\ref{img:Architecture}-C), \eqref{eq:LIF_approx_u} becomes:

\begin{equation}
 \mathbf{u}[n+1] =  \beta \mathbf{u}[n] + (1-\beta)\mathbf{i}[n+1] 
\label{eq:LIF_approx_u1}
\end{equation}
This then leads to having two separate components as illustrated in Figure~\ref{img:Architecture}-B: a Leaky Integrator (LI) neuron and a non-linear spiking function. We proceed to describe these components in detail in the following sections.

\vspace{0.5em}
\subsubsection{Parallel Leaky Integrator}~

In an LI neuron, membrane potential computation does not require the use of a non-linear spiking function. The system is therefore a Linear Time Invariant (LTI) system that can be written non-sequentially as shown by \cite{aastrom2021feedback} and used for LMU parallelization in \cite{chilkuri2021parallelizing}. We proceed to demonstrate the LTI parallelization process followed by the application of this approach to the leaky integrator component of the SPSN.

Suppose an LTI system described in \eqref{eq:LTI1} with state vector \(\mathbf{m}\), input vector \(\mathbf{v}\), and with state and input scalars \(a\) and \(b\) respectively:

\begin{equation}
\mathbf{m}[n]=a\cdot\mathbf{m}[n-1] + b\cdot\mathbf{v}[n]
\label{eq:LTI1}
\end{equation}
This can be written alternatively as,

\begin{equation}
\mathbf{m}[n]= \sum_{j=1}^{n} a^{n-j} b\cdot\mathbf{v}[j]
\label{eq:LTI2}
\end{equation}
Defining \(\mathbf{h} = [a^{n-1}b, a^{n-2}b,  ..., a^{0}b]\), we can formulate \eqref{eq:LTI2} in a non-sequential fashion as a vector multiplication:

\begin{equation}
\mathbf{m}[n] = \mathbf{h}\cdot\mathbf{v}
\label{eq:LTI3}
\end{equation}
Equation \eqref{eq:LTI3} therefore computes the state of \(\mathbf{m}\) at time \(n\). For all states, a convolution between \(\mathbf{h}\) and \(\mathbf{v}\) can be computed:

\begin{equation}
\mathbf{m}[1:n] = \mathbf{h} \ast \mathbf{v}
\label{eq:LTI4}
\end{equation}
Leveraging the convolution theorem, \eqref{eq:LTI4} can then be computed efficiently in the Fourier domain, where \(\mathcal{F}\) is the Fourier transform:

\begin{equation}
\mathbf{m}[1:n] = \mathcal{F}^{-1} \{\mathcal{F}\{\mathbf{h}\} \cdot \mathcal{F}\{\mathbf{v}\} \}
\label{eq:LTI5}
\end{equation}
In the previous equations, we have shown how a sequential LTI system \eqref{eq:LTI1} can be parallelized \eqref{eq:LTI5}. We can therefore apply this process to the LI neuron equations. We use \eqref{eq:LIF_approx_i} and \eqref{eq:LIF_approx_u1} to write respectively:

\begin{equation}
 \mathbf{i}[1:n] = \mathcal{F}^{-1} \{\mathcal{F}\{\boldsymbol{\ell}\} \cdot \mathcal{F}\{\mathbf{x}\} \}
\label{eq:LI_parall_i}
\end{equation}
\begin{equation}
 \mathbf{u}[1:n] = \mathcal{F}^{-1} \{\mathcal{F}\{\mathbf{k}\} \cdot \mathcal{F}\{\mathbf{i}[1:n]\} \}
\label{eq:LI_parall_u}
\end{equation}
where \(\boldsymbol{\ell} = [\alpha^{n-1}, \quad \alpha^{n-2},\quad ..., \quad \alpha^{0}]\) and \(\mathbf{k} = [\beta^{n-1}(1-\beta), \quad \beta^{n-2}(1-\beta), \quad ..., \quad \beta^{0}(1-\beta)]\). Finally, \eqref{eq:LI_parall_i} and \eqref{eq:LI_parall_u} can be combined as follows,

\begin{equation}
 \mathbf{u}[1:n] = \mathcal{F}^{-1} \{\mathcal{F}\{\mathbf{k}\} \cdot \mathcal{F}\{\boldsymbol{\ell}\} \cdot \mathcal{F}\{\mathbf{x}\} \}
\label{eq:LI_parall}
\end{equation}
According to \eqref{eq:LI_parall}, the LI membrane potential can be calculated in a fast parallel way. We proceed to present the spike generation phase of the proposed SPSN.

\vspace{0.5em}
\subsubsection{Stochastic Firing}~

The stochastic firing process is inspired by \cite{maass2015spike}. The author proposed an instantaneous firing probability function based on the membrane potential to approximate the experimentally observed stochastic spike generation of a biological neuron. This function is described in \eqref{eq:maass} where \(a, b, c\) are parameters to fit.
\begin{equation}
 \rho(t) = \frac{1}{a} exp\left(\frac{\mathbf{u}(t)-b}{c}\right)
\label{eq:maass}
\end{equation}
A more general mechanism was presented by \cite{gerstner2014neuronal} as an escape noise model. In this model, the strict firing threshold is replaced by a stochastic firing criterion and spikes are generated according to a probability density expressed as \( \rho(t) = f( \mathbf{u}(t) - u_{th} ) \) where \(f\) is the escape function.

In this work, we explore two stochastic spiking functions that can be applied independently at each point in time.
Given that the task consists of encoding the membrane potential into spikes, we note that various other stochastic or deterministic methods could be explored to achieve this as well.

We refer to the first solution to generate spikes from the membrane potential as the \emph{Sigmoid-Bernoulli} method. As shown in \eqref{eq:sig}, we first apply a sigmoid function \(\sigma\) to the membrane potential \(u[n]\), resulting in a spiking probability \(\rho[n]\). The spikes \(s[n]\) are then generated given these spiking probabilities using a Bernoulli distribution \eqref{eq:bern} as illustrated in Figure~\ref{img:Architecture}-C. However, this leads to gradient calculation problem because the sampling process is not differentiable. To overcome this, the \emph{Straight-Through Estimator} was proposed in \cite{bengio2013estimating} where the gradient is approximated as the identity function. We use a variant of this approach using the probability as the gradient \eqref{eq:sig_bern_grad}. During backpropagation, the descending gradient is thus multiplied by the spiking probability \(\rho[n]\). In the following, we refer to this neuron as ``Stochastic Parallelizable Spiking Neuron - Sigmoid-Bernoulli" (SPSN-SB).

\begin{equation}
 \mathbf{\rho}[n] = \sigma(\mathbf{u}[n])
\label{eq:sig}
\end{equation}
\begin{equation}
\mathbf{s}[n] = Bernoulli(\mathbf{\rho}[n])
\label{eq:bern}
\end{equation}
\begin{equation}
 \frac{\partial \mathbf{s}[n]}{\mathbf{\rho}[n]} \approx \mathbf{\rho}[n]
\label{eq:sig_bern_grad}
\end{equation}

The second stochastic spiking solution we present is the \emph{Gumbel-Softmax} method \cite{jang2016categorical}. This approach provides an efficient gradient estimator that replaces the non-differentiable sampling of a categorical random variable with a differentiable sample. It also works on Bernoulli variables. The gradient issue is then handled in this case and the use of sigmoid function is unnecessary. The spikes generated are shown in Figure~\ref{img:Architecture}-C. In the following, we refer to this neuron as ``Stochastic and Parallelizable Spiking Neuron - Gumbel Softmax" (SPSN-GS).

\section{Experiments}\label{sec:experiments}
In this Section, we present experiments to compare the proposed Stochastic Parallelizable Spiking Neuron with LIF neurons in a neuromorphic spoken digits classification task.
The following metrics will be used for comparison: a) \emph{Classification accuracy} defined the number of correct predictions divided by the total number of predictions, b) \emph{Training duration} defined time taken to train the network over the entire training set once (all networks are trained for the same number of epochs), and c) \emph{Spike frequency} defined as the mean number of spikes generated in the network per millisecond.

\subsection{Spiking Neural Network}
\label{sec:snn}
In the experiments presented below, we use a feedforward SNN with 3 fully connected hidden layers with 128 neurons in each layer. The readout layer comprises 20 LI neurons (equal to the number of classes) that do not spike. 

\begin{figure}[htbp]
\centerline{\includegraphics[width=1.0\columnwidth]{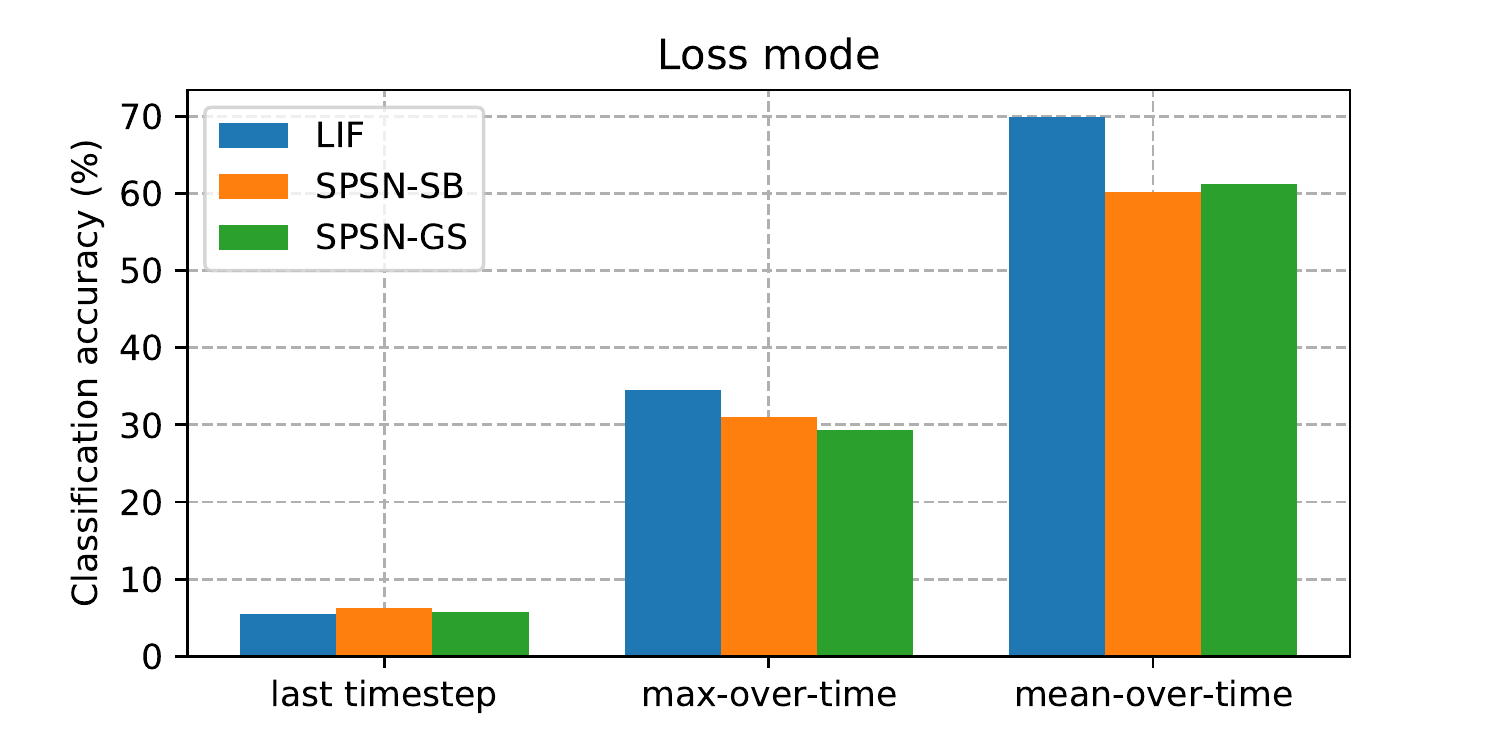}}
\caption{Comparison of classification accuracy of the three loss modes. Given the membrane potential of the readout neurons, the loss is computed based on either the last value (last time-step \cite{cramer2020heidelberg}), the max value (max-over-time \cite{cramer2020heidelberg}) or the mean value (mean-over-time \cite{yin2020effective}). These results were obtained using a single hidden layer network with \(\tau _{syn}=0.001 s\) and \(\tau _{mem}=0.001 s\).}
\label{img:loss_mode}
\end{figure}

We use the cross entropy loss function defined in \eqref{eq:loss} where \(C\) is the number of classes, \(\mathbbm{1}\) is the indicator function, \(y\) is the target class, \(mean\) is the mean-over-time function applied to the membrane potential \(u_i\) of neuron \(i\). The mean-over-time function was chosen as it resulted in the best accuracy in preliminary experiments comparing the performance of the mean-over-time \cite{yin2020effective}, max-over-time \cite{cramer2020heidelberg}, and last time step \cite{cramer2020heidelberg} approaches (see Figure~\ref{img:loss_mode}). 
Note that the default parameter values used during this preliminary experiment differ from the final values used in subsequent experiments where a thorough optimization was performed.

\begin{equation}
\mathcal{L} = - \sum_{i=1}^{C} \mathbbm{1}(i=y) \cdot log \left[ \frac{exp(mean(u_i)) }{ \sum_{j=1}^{C} exp(mean(u_j))} \right ]
\label{eq:loss}
\end{equation}

All parameter values are summarized in Table \ref{table:parameters}.

\begin{figure*}[htbp]
\centerline{\includegraphics[width=1.0\textwidth]{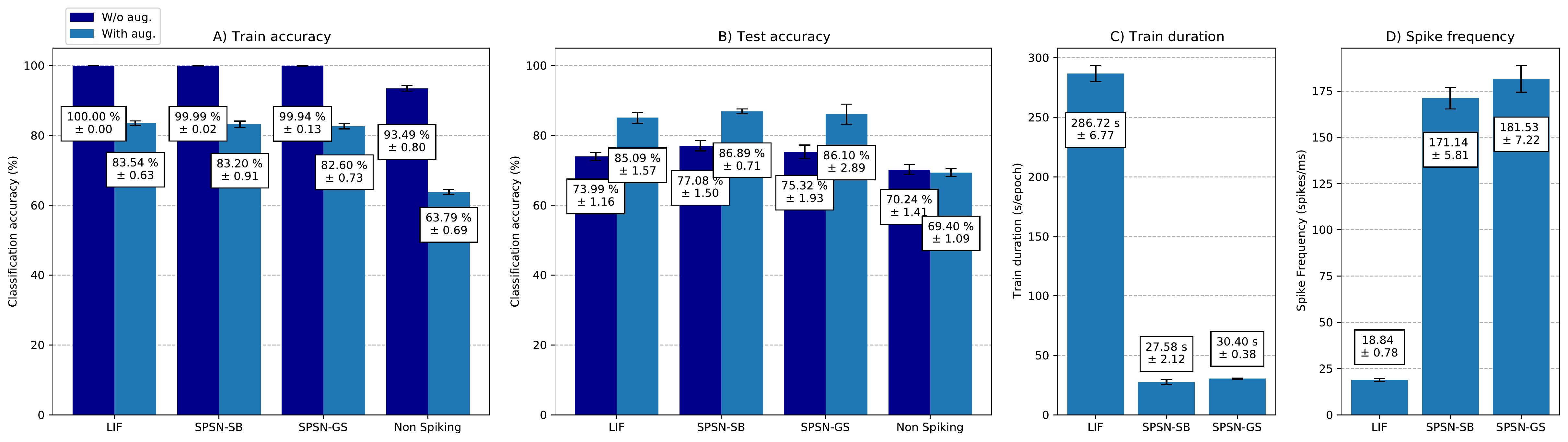}}

\caption{Experimental results. A) Train set classification accuracy for the 4 networks under test, with and without data augmentation. B) Test set classification accuracy, with and without data augmentation. C) Training duration for the SNN networks presented as the mean duration for 1 training epoch. D) Spike frequencies for the SSN models, defined as the mean number of spikes in the network per millisecond.}

\label{img:results}
\end{figure*}

\begin{table}
\setlength{\tabcolsep}{15pt}
\centering
\caption{Parameters Used in Experiments}
\label{table:parameters}
\begin{tabular}{|ll|} 
\hline
\textbf{Parameters} & \textbf{Values}  \\ 
\hline
\# hidden layers & 3       \\ 
\hline
\# classes       & 20      \\ 
\hline
\# epochs        & 200     \\ 
\hline
 \(\tau _{mem}\) &  0.02 s       \\
\hline
 \(\tau _{syn}\) &  0.02 s     \\
\hline
 \(u_{th}\) &  1 volt      \\
\hline
 \(\Delta_t\) &  0.001 s     \\
\hline
 \(k_{shift}\) &  0.1     \\
\hline
 \(k_{scale}\) &  0.3     \\
\hline
  batch size &  64     \\
\hline
  loss function &  cross entropy     \\
\hline
  optimizer &  Adamax     \\
\hline
  learning rate &  0.001     \\
\hline
  \(\theta_{reg}\) (SPSN-SB) &  0.4     \\
\hline
  \(\theta_{reg}\) (SPSN-GS) &  0.1     \\
\hline
\end{tabular}
\end{table}

\subsection{Dataset}
We use the Spiking Heidelberg Digits (SHD) dataset for our experiments \cite{cramer2020heidelberg}. It consists of approximately 10\,000 studio recordings of spoken digits from 0 to 9 in both German and English language (20 classes). Each sample is a set of spike trains from 700 input channels generated by an artificial cochlea model \cite{cramer2020heidelberg}.

As shown by SHD authors, data augmentation has a positive impact on generalization. Three methods of data augmentation were proposed in \cite{nowotny2022loss}, however only the ``random shift augmentation" method was shown to improve generalization noticeably. In addition to this approach, we propose an additional method ``random scale augmentation" which preliminary experiments demonstrated to be effective. We therefore use these two augmentation methods in our experiments, defined as follows:

\begin{itemize}
    \item \textit{Random shift augmentation:} Input spikes are shifted by a random factor along the channel axis. The factor is chosen uniformly in the interval [\(-k_{shift}\); \quad \(k_{shift}\)] and rounded to the nearest integer.
    \item \textit{Random scale augmentation:} The input spikes are scaled along the two axes by a random factor uniformly chosen in the interval [\((1-k_{scale})\); \quad \((1+k_{scale})\)], followed by a nearest neighbour interpolation. This is conceptually similar to a zoom applied to an image. 
\end{itemize}

\subsection{Spike Frequency Regularization}
As can be seen in Figure~\ref{img:Architecture}-C, the spike frequency of SPSN is significantly higher than LIF. Contrary to LIF which only emits a spike when the membrane potential is higher than the threshold, SPSN emits spikes according to a spiking probability computed from the membrane potential. Therefore, whenever this probability is non-zero, a spike may occur. However, if we were to reduce the network weights and biases, the membrane potentials would have a lower amplitudes, thus impacting the spiking probability and reducing the spiking frequency.

To limit the spike frequency in the network, we add a regularization term to the loss function \(\mathcal{L}\). A similar regularization was proposed in \cite{cramer2020heidelberg}. This term uses a maximum threshold to limit the average frequency of neuron spikes. The regularization formula is defined in \eqref{eq:regularization}, where \(N\) is the total number of neurons in the network (excluding the readout neurons as they do not spike), \(T\) is the number of time-steps and \(S_n[t]\) is the presence or absence of spike at time \(t\) for neuron \(n\).

\begin{equation}
 \mathcal{L}_{reg} = relu \left [ \sum_{n=1}^{N} \left(\frac{1}{T} \sum_{t=1}^{T} S_n[t] \right)- \left(\theta_{reg}*N \right) \right ]^2
\label{eq:regularization}
\end{equation}

\section{Results}\label{sec:results}
To compare the proposed SPSN neurons with LIFs, we built 3 SNNs using the architecture defined in Section~\ref{sec:snn}, one for each neuron type: LIF, SPSN-SB, and SPSN-GS. We also compare the proposed approach with a non-spiking network using the same architecture defined in Section~\ref{sec:snn} using Rectified Linear Unit (ReLU) activation functions. We trained and tested each network using the SHD dataset and recorded the resulting performance. Experiments were repeated 10 times and the average results are reported in Figure~\ref{img:results}.

In Figure~\ref{img:results}-A, we note that the accuracy on the train set when data augmentation is not applied reaches 100\%. This is caused by over-fitting. However, when data augmentation is applied, the train accuracy is considerably reduced. This is observed for all networks.

In Figure~\ref{img:results}-B, the effect of data augmentation becomes clear: the accuracy on the test set is increased by approximately 10\%. The classification accuracy increases from 73.99\% to 85.09\% for LIF, from 77.08\% to 86.89\% for SPSN-SB and from 75.32\% to 86.1\% for SPSN-GS. The accuracy is quite similar for the three neuron types, with SPSN being slightly better. We note that with data augmentation, the test set accuracy is even greater than the train set accuracy.
However, the non-spiking network yields the lowest accuracy and data augmentation does not lead to an improvement.

In Figure~\ref{img:results}-C, we present the average training duration for 1 epoch (with data augmentation). We observe 286.72 seconds for LIF, 27.58 seconds for SPSN-SB, and 30.4 seconds for SPSN-GS. The proposed stochastic neurons are therefore approximately 10 times faster than LIF.

Similarly, in Figure~\ref{img:results}-D, we plot the average spiking frequency for the three neuron types. The frequency is the average number of spikes in the network at each millisecond. The frequency is 18.84 spikes/ms for LIF, 171.14 spikes/ms for SPSN-SB and 181.53 spikes/ms for SPSN-GS. The SPSN therefore generated around 10 times more spikes than LIF.

\begin{figure}[htbp]
\centerline{\includegraphics[width=0.9\columnwidth]{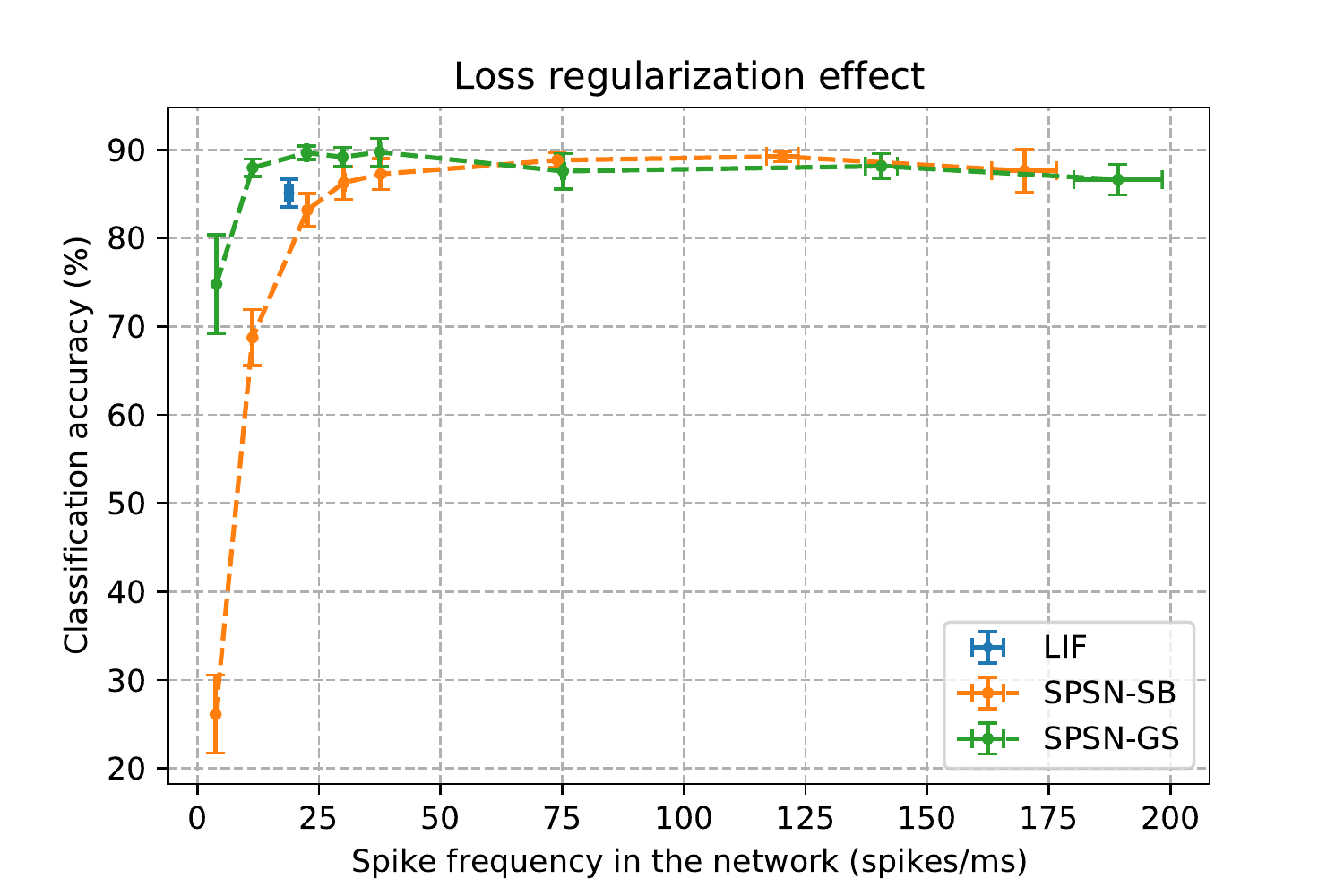}}
\caption{Impact of spike frequency regularization on classification accuracy. The regularization is applied to the SPSN to reduce the spike frequency in the network. The experiments are performed 10 times for each regularization threshold \(\theta_{reg}\) and the threshold vary in the range [0.01, 0.03, 0.06, 0.08, 0.1, 0.2, 0.4, 0.6].}
\label{img:regularization}
\end{figure}

To study the effect of spiking frequency regularization on stochastic neurons, we varied the regularization threshold \(\theta_{reg}\) in the range [0.01, 0.03, 0.06, 0.08, 0.1, 0.2, 0.4, 0.6]. We then plot the accuracy evolution curve as a function of the average spiking frequency in the network. As seen in Figure~\ref{img:regularization}, the precision drops when the spiking frequency is less than approximately 25 spikes/ms and is constant for spiking frequencies greater than 50 spikes/ms. At around 25 spikes/ms, we note that SPSN-GS outperforms LIF with a similar spiking frequency, yielding even better results than without regularization. Note that the regularization does not impact the training duration. The best accuracies achieved are 89.23\% (120 spikes/s with \(\theta_{reg}\)=0.4) for SPSN-SB and 89.73\% (37 spikes/s with \(\theta_{reg}\)=0.1) for SPSN-GS.

\section{Discussion}\label{sec:discussion}

\subsection{Classification Performance}
The overall classification accuracies are summarized in Table~\ref{table:accuracy}. When no data augmentation is applied, SPSN outperforms LIF. This is further confirmed when data are augmented bringing a significant increase in test accuracy of approximately 10\%. 

The non-spiking network performs worse than the spiking networks. This is likely due to the inherent ability of spiking neurons to process temporal information, communicating with each other through binary values, unlike the non-spiking network which communicates in real values. The non-spiking network would likely yield better results with a recurrent architectures.

In addition to that, the results in Figure~\ref{img:regularization} demonstrate that the high spiking frequency issue of SPSN can be solved effectively. The regularization applied allows for a considerable reduction in the spiking frequency without decreasing classification accuracy. For a similar spike frequency, SPSN-GS clearly outperforms LIF. Regularization even results in increased test accuracy for certain spike frequency values, achieving better results for SPSN-SB (89.23\%) and SPSN-GS (89.73\%).

According to the leader board presented by SHD dataset authors\footnote{\url{https://zenkelab.org/resources/spiking-heidelberg-datasets-shd/}}, with a simpler network, SPSN achieves only 3\% lower performance compared to state-of-the-art for a feedforward network (92.4\% by \cite{yu2022stsc}) and 4\% higher than the baseline accuracy (83.2\% with a recurrent network \cite{cramer2020heidelberg}).

\begin{table}
\setlength{\tabcolsep}{4pt}
\centering
\caption{Classification accuracy summarized}
\label{table:accuracy}
\begin{tabular}{|l|lll|}
\hline
\multirow{2}{*}{\textbf{Model}} & \multicolumn{3}{c|}{\textbf{Test Accuracy (\%)}}                                                                                       \\ \cline{2-4} 
                                & \multicolumn{1}{c}{\textbf{Basic}} & \multicolumn{1}{c}{\textbf{Data aug.}} & \multicolumn{1}{c|}{\textbf{Data aug. + Regularization}} \\ \hline
\textbf{LIF}                    & 73.99 ± 1.16                       & 85.09 ± 1.57                           & -                                                        \\ \hline
\textbf{SPSN-SB}                & \textbf{77.08 ± 1.5}               & \textbf{86.89 ± 0.71}                  & 89.23 ± 0.57                                             \\ \hline
\textbf{SPSN-GS}                & 75.32 ± 1.93                       & 86.10 ± 2.89                           & \textbf{89.73 ± 1.57}                                    \\ \hline
\textbf{Non spiking}     & 70.24 ± 1.41                       & 69.40 ± 1.09                           & -                                    \\ \hline
\end{tabular}
\end{table}

\subsection{Speed up}
\begin{figure}[htbp]
\centerline{\includegraphics[scale=0.55]{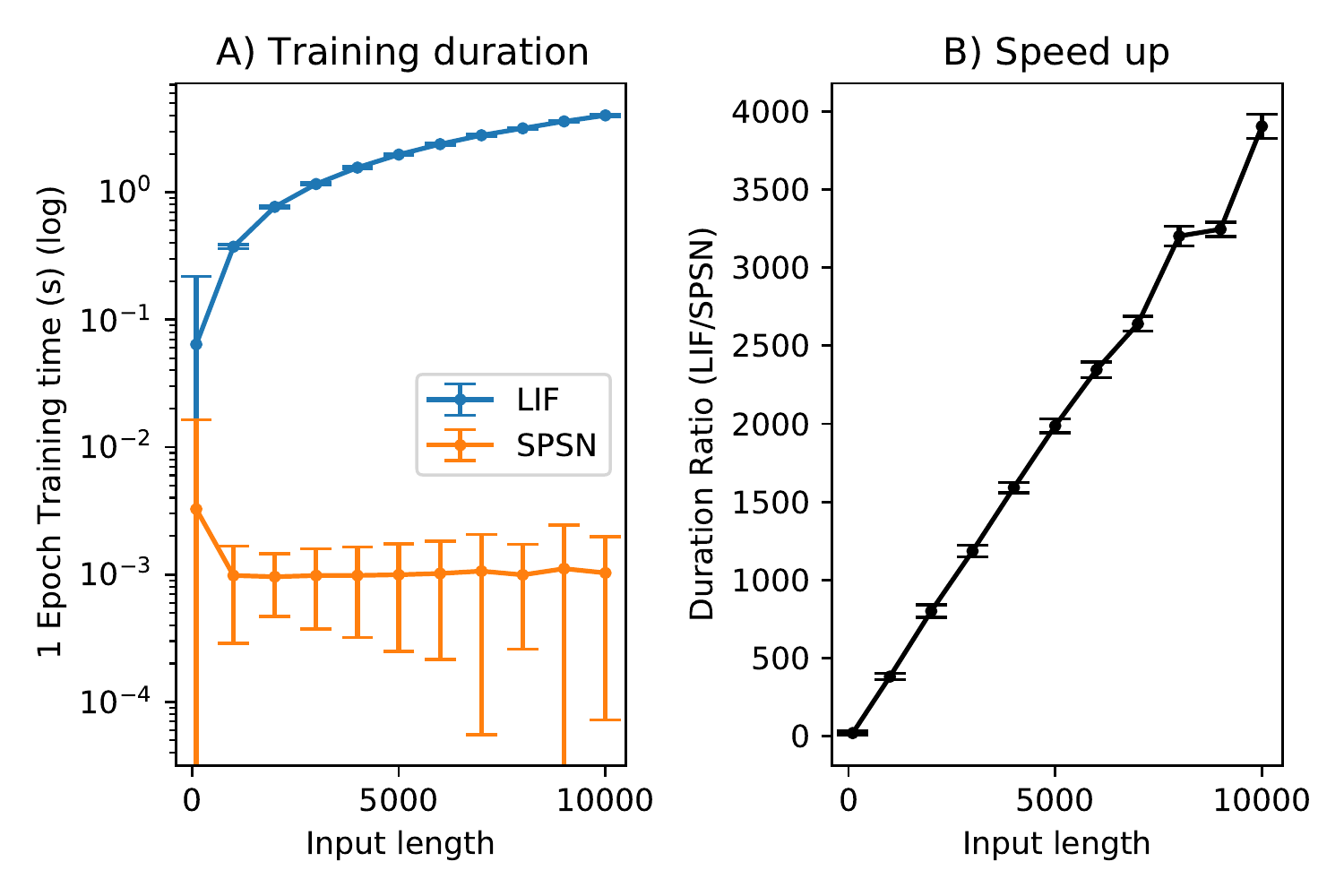}}
\caption{Simulation of a training process (Forward+Backward) for different input sequence lengths. A) The left subplot shows the duration of training as a function of input sequence length for LIF and SPSN-SB.
Note that the SPSN error bars appear larger due to the logarithmic scale.
B) The right subplot shows the ratio between the LIF and SPSN training durations, providing the speed-up value. These experiments were run on a single NVIDIA Tesla V100-SXM2 GPU, 10 times for each input length.}
\label{img:time_comparison}
\end{figure}

In Figure~\ref{img:results}-C, it is shown that SPSN is 10 times faster than LIF for the SHD dataset. We conducted additional experiments to determine the speedup of SPSN as a function of input sequence length by simulating a training process (forward + backward) on a one layer network. The input is a random sequence with a length fixed in the range [100, 10000]. We repeat this 10 times for each sequence length and plot the training duration as a function of input sequence length for LIF and SPSN-SB in Figure~\ref{img:time_comparison}. We note that training duration increases with increasing sequence length for LIF, while for SPSN the training duration is constant. This is due to the sequential computation for LIF and the parallel computation for SPSN. By calculating the ratio between LIF and SPSN training durations, we found that the speed up factor reaches up to 4000 times for an input sequence length of 10\,000.

\subsection{Prediction robustness to stochasticity}
Due to the stochasticity of the SPSN, the issue of robustness of the predictions can be raised. Indeed, for a trained network with frozen weights, a given input can generate different outputs for different inference trials. To test this robustness, we pass the same input 10 times and observe the readout neurons' output over time. For each readout neuron, we then plot the mean over-trials and the standard deviation to see the effect of stochasticity on network prediction.

\begin{figure}[htbp]
\centerline{\includegraphics[scale=0.4]{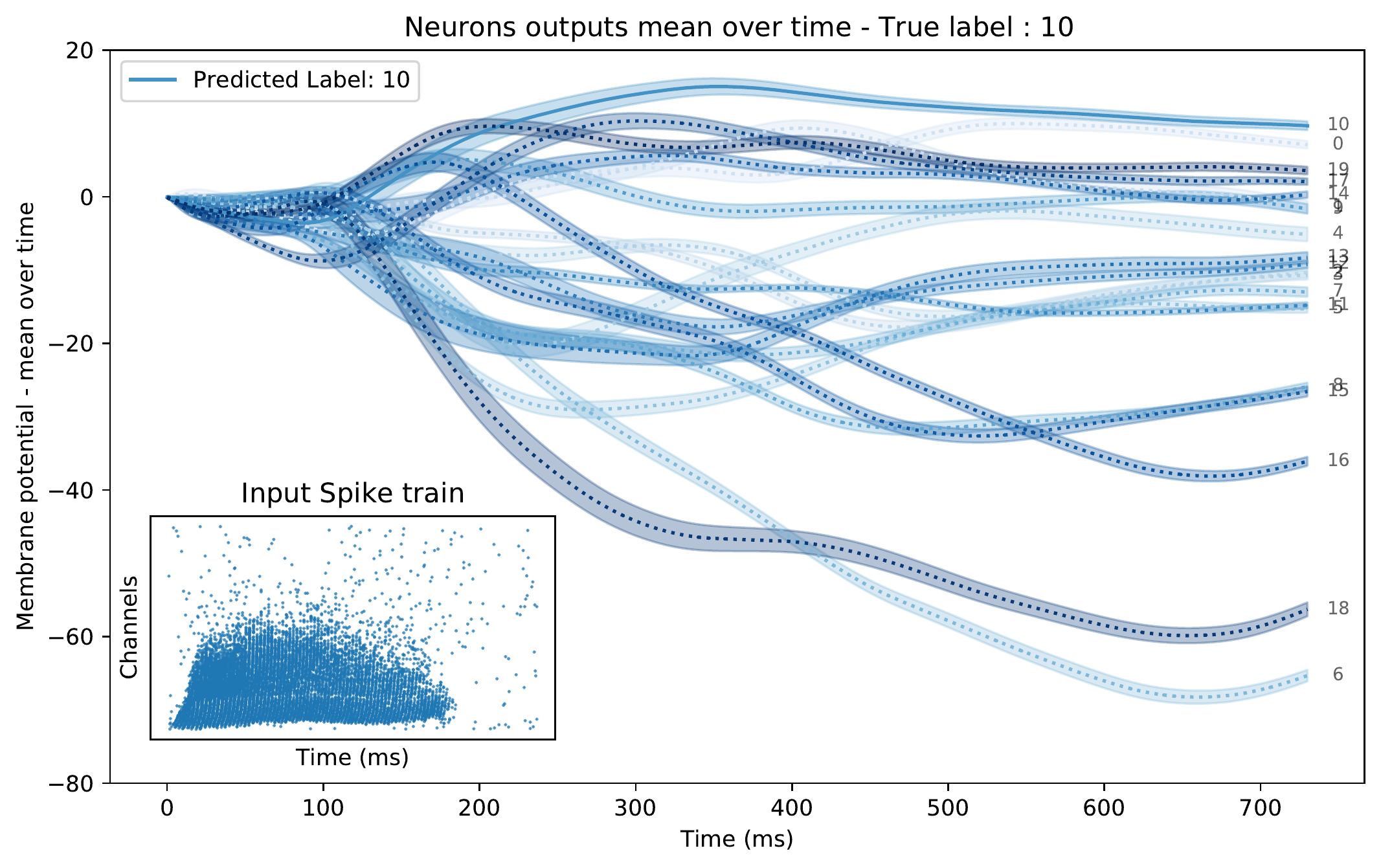}}
\caption{Analysis of SPSN prediction robustness to network stochasticity. The same input is introduced 10 times in a trained network and the 20 output neurons' membrane potentials are observed as the mean over time of the membrane potential. Because of stochasticity, each trial gives a different output, however the standard deviations demonstrate that in each trial the correct label is predicted.}
\label{img:neuron_output}
\end{figure}

As we can observe in Figure~\ref{img:neuron_output}, the standard deviation is small enough to lead to the same label prediction in each trial. Despite the stochasticity, therefore, the network dynamic doesn't change enough to result in a change in prediction. The SPSN can therefore be considered reliable and robust to stochasticity.

\subsection{SPSN-SB vs SPSN-GS}

The results of the previous experiments demonstrate that both SPSN-SB and SPSN-GS achieve comparable performance levels. However, a detailed analysis of the experimental results presented in Table~\ref{table:accuracy} reveals that SPSN-SB exhibits a slight improvement in performance, both with and without data augmentation, whereas SPSN-GS performs better when spike regularization is applied. Furthermore, the findings from Figure~\ref{img:regularization} provide evidence that SPSN-GS demonstrates improved tolerance to low spike frequencies. From a conceptual standpoint, SPSN-SB is advantageous due to its simplicity of implementation, while SPSN-GS offers the advantage of overcoming the gradient propagation issue via a reparametrization trick.

\subsection{Limitations \& Future work}

Despite the interesting results in terms of training acceleration and classification performance, there are some limitations of the proposed SPSN in its current state. The primary limitation is related to available network architectures. Indeed, the SPSN is not well-suited for application in a recurrent SNN as it will no longer be parallelizable. The recurrence in the network would in fact re-introduce the sequential simulation that SPSN sought to overcome. The second limitation is about the generalization of the presented results. In fact only one dataset has been used during experiments. These issues will be addressed in future work to enable recurrent architectures for the SPSN while using diverse datasets for experiments. In addition, an analysis will be conducted on the characteristics of spike trains generated by SPSN to extend its comparison with LIF neurons.

\section{Conclusion}\label{sec:conclusion}
In this work, we presented a new spiking neuron for fast SNN training. By taking advantage of the ability of LTI systems to be parallelized over time, the proposed Stochastic Parallelizable Spiking Neuron (SPSN) can be simulated in a parallel fashion, significantly reducing training time on GPUs. The SPSN relies on stochastic spiking functions to convert membrane potential to spikes. A comparison of feedforward SPSN and LIF networks on the Spiking Heidelberg Digits (SHD) dataset demonstrated that SPSN achieves a higher classification accuracy while being 10 times faster to train. The proposed approach also outperformed non-spiking networks with the same network architecture. Two data augmentation methods were presented that increase classification accuracy by approximately 10\%. A regularization method was proposed to overcome the high spiking frequency of the SPSN, resulting in higher accuracy than LIF for a similar spiking frequency. The regularization method also increased classification accuracy, achieving only 3\% less than state-of-the-art performance for a similar architecture on the SHD dataset. For longer input sequences of 10\,000 time-steps, the SPSN resulted in 4000 times faster training, thus demonstrating the potential of the proposed approach to accelerate SNN training for very large datasets.

\bibliographystyle{IEEEtran}
\bibliography{sample}

\end{document}